\documentclass[final]{cvpr}
\usepackage{tabulary,xspace}
\usepackage{tabularx}
\usepackage{times}
\usepackage{epsfig}
\usepackage{graphicx}
\usepackage{fixmath,mathtools,nicefrac,mmstyle}
\usepackage[table,xcdraw]{xcolor}
\usepackage{amsmath}
\usepackage{amssymb}
\usepackage{caption}
\usepackage{multirow}
\usepackage{booktabs}
\usepackage[section]{placeins}
\usepackage{float}
\usepackage{wrapfig}

\usepackage[pagebackref,breaklinks,colorlinks]{hyperref}

\definecolor{ourscolor}{gray}{.9}
\newcommand{\ours}[1]{\cellcolor{ourscolor}{#1}}
\newcommand{\authorskip}{\hspace{2.5mm}}

\def\cvprPaperID{****} 
\def\confName{CVPR}
\def\confYear{2021}
\graphicspath{{./figs/}{./figs/result/}}

\begin{document}

\title{RTMPose: Real-Time Multi-Person Pose Estimation based on MMPose}

\author{Tao Jiang \authorskip Peng Lu \authorskip Li Zhang \authorskip Ningsheng Ma \authorskip
Rui Han \authorskip \\ Chengqi Lyu \authorskip Yining Li \authorskip Kai Chen \\[2mm]
Shanghai AI Laboratory \\
{\tt\small $\left \{\text{jiangtao, lupeng, zhangli, maningsheng, hanrui, lvchengqi, liyining, chenkai}\right\}$@pjlab.org.cn} \\
}

\maketitle


\begin{abstract}
   Recent studies on 2D pose estimation have achieved excellent performance on public benchmarks, yet its application in the industrial community still suffers from heavy model parameters and high latency.
   To bridge this gap, we empirically explore key factors in pose estimation including paradigm, model architecture, training strategy, and deployment, and present a high-performance real-time multi-person pose estimation framework, \textbf{RTMPose}, based on MMPose.
   Our RTMPose-m achieves \textbf{75.8\% AP on COCO} with \textbf{90+ FPS} on an Intel i7-11700 CPU and \textbf{430+ FPS} on an NVIDIA GTX 1660 Ti GPU, and RTMPose-x achieves \textbf{65.3\% AP on COCO-WholeBody}.
   To further evaluate RTMPose's capability in critical real-time applications, we also report the performance after deploying on the mobile device. Our RTMPose-s model achieves \textbf{72.2\% AP on COCO} with \textbf{70+ FPS} on a Snapdragon 865 chip, outperforming existing open-source libraries. 
   Our code and models are available at \url{https://github.com/open-mmlab/mmpose/tree/main/projects/rtmpose}.

\end{abstract}

\section{Introduction}\label{sec:Introduction}

Real-time human pose estimation is appealing to various applications such as human-computer interaction, action recognition, sports analysis, and VTuber techniques. Despite the stunning progress \cite{SunXLW19, xu2022vitpose} on academic benchmarks \cite{lin2014coco, li2018crowdpose}, it remains a challenging task to perform robust and real-time multi-person pose estimation on devices with limited computing power. Recent attempts narrow the gap with efficient network architecture \cite{Yulitehrnet21, blazepose, RonnyVotel2023NextGenerationPD} and detection-free paradigms \cite{kreiss2019pifpaf, dekr, shi2022end}, which is yet inadequate to reach satisfactory performance for industrial applications.

\begin{figure}[ht]
    \centering
    \includegraphics[width=\linewidth]{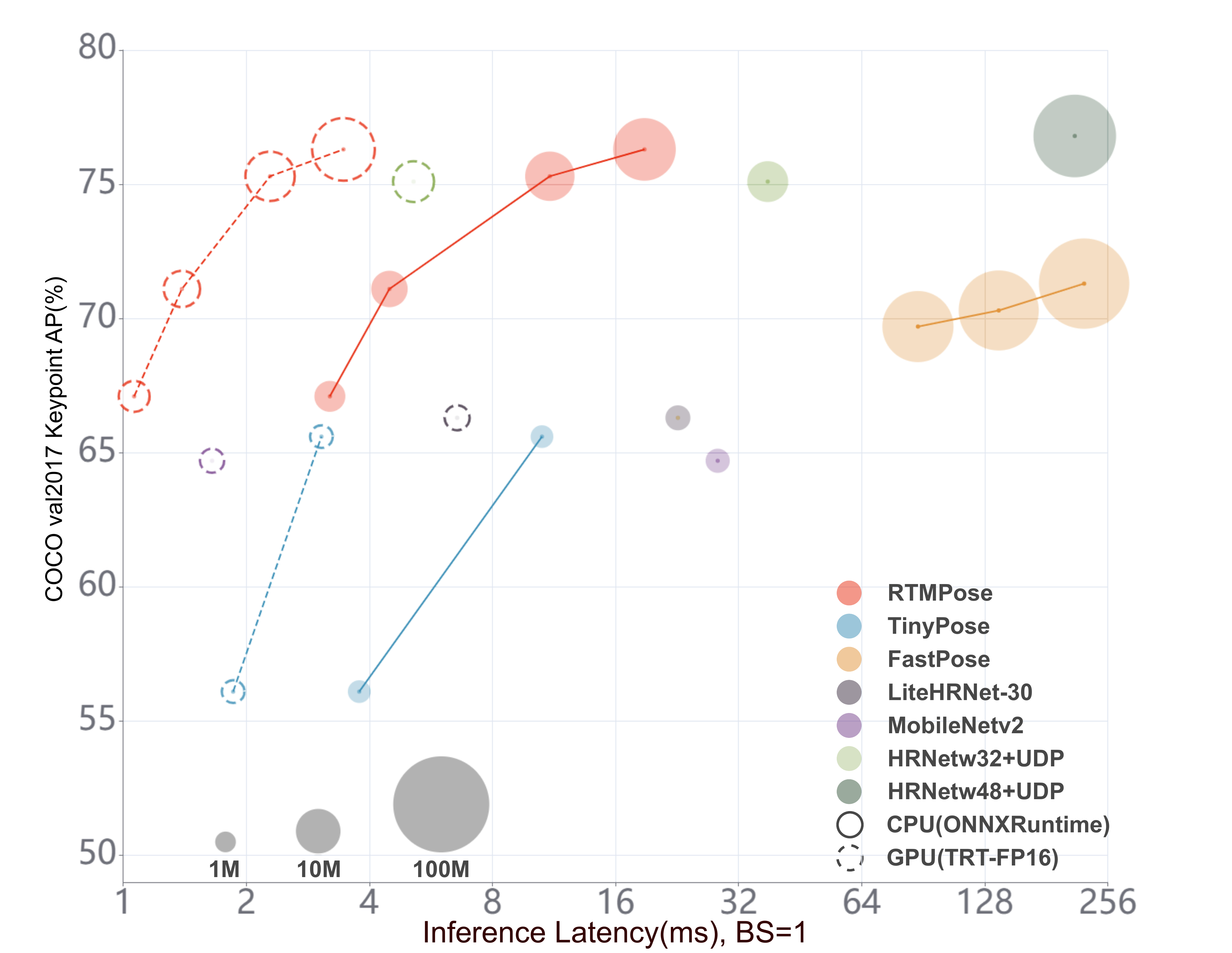}
    \caption{Comparison of RTMPose and open-source libraries on COCO val set regarding model size, latency, and precision. The circle size represents the relative size of model parameters.}
    \label{fig:compare_latency}
\end{figure}

In this work, we empirically study key factors that affect the performance and latency of 2D multi-person pose estimation frameworks from five aspects: paradigm, backbone network, localization method, training strategy, and deployment. With a collection of optimizations, we introduce \textbf{RTMPose}, a new series of \textbf{R}eal-\textbf{T}ime \textbf{M}odels for \textbf{Pose} estimation.

First, RTMPose employs a top-down approach by using an off-the-shelf detector to obtain bounding boxes and then estimating the pose of each person individually. Top-down algorithms have been stereotyped as accurate but slow, due to the extra detection process and increasing workload in crowd scenes. However, benefiting from the excellent efficiency of real-time detectors \cite{nanodet, lyu2022rtmdet}, the detection part is no longer a bottleneck of the inference speed of top-down methods. In most scenarios (within 6 persons per image), the proposed lightweight pose estimation network is able to perform multiple forward passes for all instances in real time.

Second, RTMPose adopts CSPNeXt\cite{lyu2022rtmdet} as the backbone, which is first designed for object detection. Backbones designed for image classification \cite{He2015, sandler2018mobilenetv2} are suboptimal for dense prediction tasks like object detection, pose estimation and semantic segmentation, etc. Some backbones leveraging high-resolution feature maps \cite{SunXLW19, Yulitehrnet21} or advanced transformer architectures \cite{dosovitskiy2020vit} achieve high accuracy on public pose estimation benchmarks, but suffer from high computational cost, high inference latency, or difficulties in deployment. CSPNeXt shows a good balance of speed and accuracy and is deployment-friendly.

Third, RTMPose predicts keypoints using a SimCC-based\cite{SimCC} algorithm that treats keypoint localization as a classification task. Compared with heatmap-based algorithms \cite{xiao2018simple, Zhang_2020_CVPR, Huang_2020_CVPR, xu2022vitpose}, the SimCC-based algorithm achieves competitive accuracy with lower computational effort. Moreover, SimCC uses a very simple architecture of two fully-connected layers for prediction, making it easy to deploy on various backends.

Fourth, we revisit the training settings in previous works \cite{blazepose, li2021human, lyu2022rtmdet}, and empirically introduce a collection of training strategies applicable to the pose estimation task. Our experiments demonstrate that this set of strategies bring significant gains to proposed RTMPose as well as other pose estimation models.

Finally, we jointly optimize the inference pipeline of the pose estimation framework. We use the skip-frame detection strategy proposed in \cite{blazepose} to reduce the latency and improve the pose-processing with pose Non-Maximum Suppression (NMS) and smoothing filtering for better robustness. In addition, we provide a series of RTMPose models with t/s/m/l/x sizes to cover different application scenarios with the optimum performance-speed trade-off.

We deploy RTMPose with different inference frameworks (PyTorch, ONNX Runtime, TensorRT, ncnn) and hardwares (i7-11700, GTX1660Ti, Snapdragon865) to test the efficiency.

As shown in Fig.~\ref{fig:compare_latency}, We evaluate the efficiency of RTMPose with various inference frameworks (PyTorch, ONNX Runtime, TensorRT, ncnn) and hardwares (Intel i7-11700, GTX 1660Ti, Snapdragon 865). Our RTMPose-m achieves 75.8\% AP (with flipping) on COCO val set with 90+ FPS on an Intel i7-11700 CPU, 430+ FPS on an NVIDIA GeForce GTX 1660 Ti GPU, and 35+ FPS on a Snapdragon 865 chip. Using the high-performance real-time object detection model RTMDet-nano in our pose estimation pipeline, RTMPose-m can achieve 73.2\% AP. With the help of MMDeploy\cite{mmdeploy}, RTMPose can also be easily deployed to various backends like RKNN, OpenVINO, PPLNN, etc.


\section{Related Work}

\begin{figure*}[t]
    \centering
    \includegraphics[width=1\textwidth]{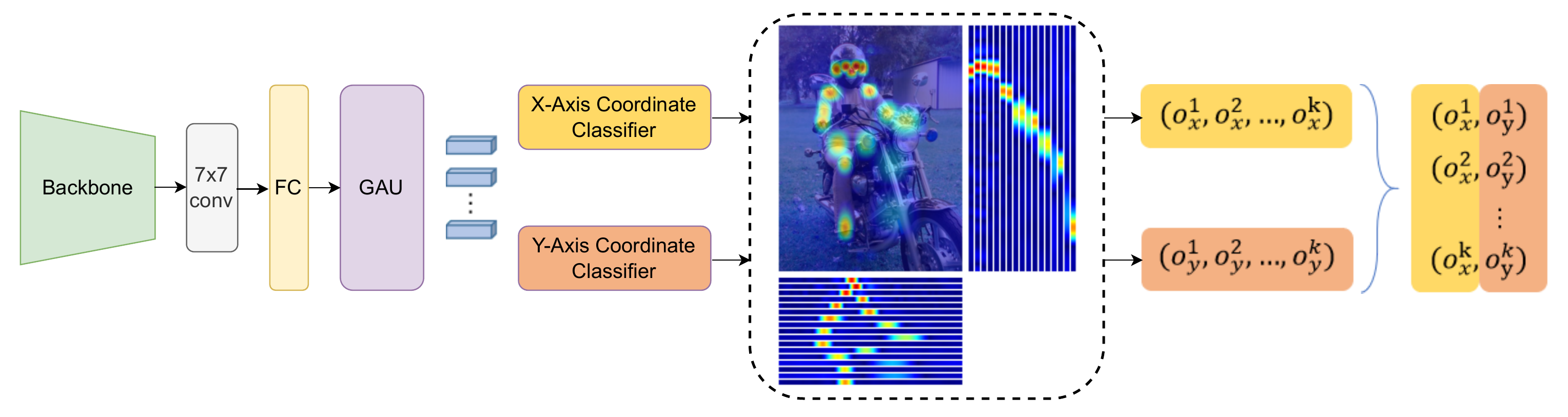}
    \caption{The overall architecture of RTMPose, which contains a convolutional layer, a fully-connected layer and a Gated Attention Unit (GAU) to refine K keypoint representations. After that 2d pose estimation is regarded as two classification tasks for x-axis and y-axis coordinates to predict the horizontal and vertical locations of keypoints.} 
    \label{fig:model}
\end{figure*}

\paragraph{Bottom-up Approaches.} 
Bottom-up algorithms ~\cite{pishchulin2016deepcut, cao2017realtime, newell2017associative, luo2021rethinking, cheng2020higherhrnet, dekr, kreiss2019pifpaf, jin2020differentiable} detect instance-agnostic keypoints in an image and partition these keypoints to obtain the human pose. The bottom-up paradigm is considered suitable for crowd scenarios because of the stable computational cost regardless the number of people increases. However, these algorithms often require a large input resolution to handle various person scales, making it challenging to reconcile accuracy and inference speed.

\paragraph{Top-down Approaches.} 
Top-down algorithms use off-the-shelf detectors to provide bounding boxes and then crop the human to a uniform scale for pose estimation. Algorithms ~\cite{xiao2018simple, cai2020learning, SunXLW19, liu2021polarized, xu2022vitpose} of the top-down paradigm have been dominating public benchmarks. The two-stage inference paradigm allows both the human detector and the pose estimator to use relatively small input resolutions, which allows them to outperform bottom-up algorithms in terms of speed and accuracy in non-extreme scenarios (i.e. when the number of people in the image is no more than 6). Additionally, most previous work has focused on achieving state-of-the-art performance on public datasets, while our work aims to design models with better speed-accuracy trade-offs to meet the needs of industrial applications.

\paragraph{Coordinate Classification.} Previous pose estimation approaches usually regard keypoint localization as either coordinate regression (e.g. ~\cite{Toshev_2014_CVPR, li2021human, mao2022poseur}) or heatmap regression (e.g. ~\cite{xiao2018simple, huang2020devil, Zhang_2020_CVPR, xu2022vitpose}). SimCC~\cite{SimCC} introduces a new scheme that formulates keypoint prediction as classification from sub-pixel bins for horizontal and vertical coordinates respectively, which brings about several advantages. First, SimCC is freed from the dependence on high-resolution heatmaps, thus allowing for a very compact architecture that requires neither high-resolution intermediate representations ~\cite{SunXLW19} nor costly upscaling layers ~\cite{xiao2018simple}. Second, SimCC flattens the final feature map for classification instead of involving global pooling ~\cite{Toshev_2014_CVPR} and therefore avoids the loss of spatial information. Third, the quantization error can be effectively alleviated by coordinate classification at the sub-pixel scale, without the need for extra refinement post-processing ~\cite{Zhang_2020_CVPR}. These qualities make SimCC attractive for building lightweight pose estimation models. In this work, we further exploit the coordinate classification scheme with optimizations on model architecture and training strategy.

\paragraph{Vision Transformers.} 
Transformer-based architectures~\cite{vaswani2017attention} ported from modern Natural Language Processing (NLP) have achieved great success in various vision tasks like representation learning ~\cite{dosovitskiy2020vit, liu2021swin}, object detection~\cite{detr, zhu2020deformable, vitdet}, semantic segmentation ~\cite{zheng2021rethinking}, video understanding~\cite{liu2022video, bertasius2021space, fan2021multiscale}, as well as pose estimation~\cite{xu2022vitpose, yang2021transpose, mao2022poseur, li2021tokenpose, shi2022end}. ViTPose~\cite{xu2022vitpose} leverages the state-of-the-art transformer backbones to boost pose estimation accuracy, while TransPose~\cite{yang2021transpose} integrates transformer encoders with CNNs to efficiently capture long-range spatial relationships. Token-based keypoint embedding is introduced to incorporate visual cue querying and anatomic constraint learning, shown effective in both heatmap-based~\cite{li2021tokenpose} and regression-based~\cite{mao2022poseur} approaches. PRTR~\cite{li2021pose} and PETR~\cite{shi2022end} propose end-to-end multi-person pose estimation frameworks with transformers, inspired by the pioneer in detection~\cite{detr}.
Previous pose estimation approaches with transformers either use a heatmap-based representation or retained both pixel tokens and keypoint tokens, which results in high computation costs and makes real-time inference difficult. In contrast, we incorporate the self-attention mechanism with a compact SimCC-based representation to capture the keypoint dependencies, which significantly reduces the computation load and allows real-time inference with advanced accuracy and efficiency.


\section{Methodology}

\begin{figure}[tbp]
    \includegraphics[width=0.5\textwidth]{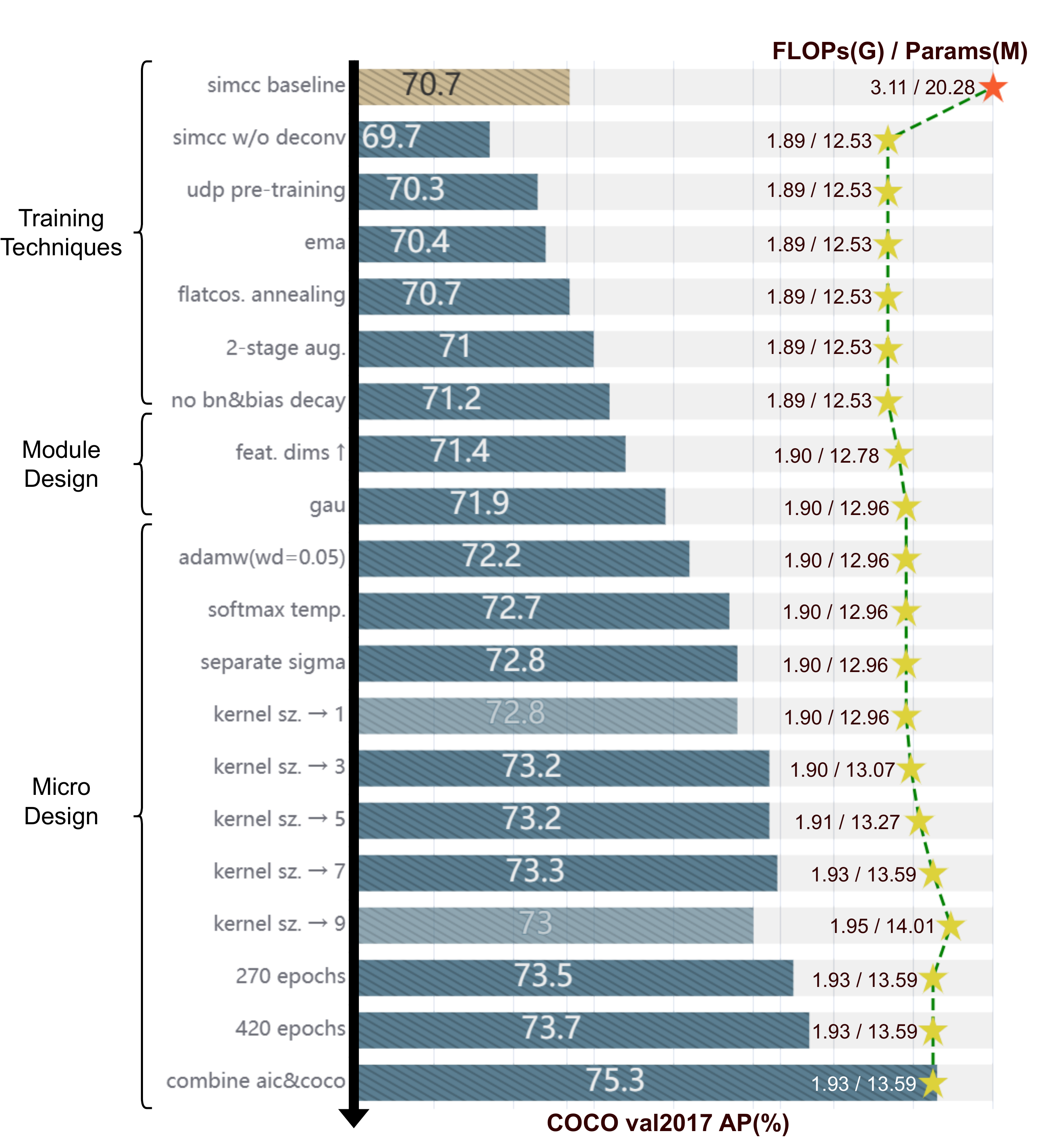}
    \caption{Step-by-step improvements from a SimCC baseline.}
    \label{fig:roadmap}
\end{figure}

In this section, we expound the roadmap we build RTMPose following the coordinate classification paradigm. We start by refitting SimCC~\cite{SimCC} with more efficient backbone architectures, which gives a lightweight yet strong baseline (\ref{sec:baseline}). We adopt the training strategies proposed in~\cite{lyu2022rtmdet} with minor tweaks to make them more effective on the pose estimation task. The model performance is further improved with a series of delicate modules (\ref{sec:module}) and micro designs (\ref{sec:micro}). Finally, we jointly optimize the entire top-down inference pipeline toward higher speed and better reliability. The final model architecture is shown in Fig.~\ref{fig:model}, and Fig.~\ref{fig:roadmap} illustrates the step-by-step gain of the roadmap.

\subsection{SimCC: A lightweight yet strong baseline}\label{sec:baseline}

\paragraph{Preliminary}
SimCC~\cite{SimCC} formulates the keypoint localization as a classification problem. The core idea is to divide the horizontal and vertical axes into equal-width numbered bins and discretize continuous coordinates into integral bin labels. Then the model is trained to predict the bin in which the keypoint is located. The quantization error can be reduced to a subpixel level by using a large number of bins.

Thanks to this novel formulation, SimCC has a very simple structure that uses a $1\times1$ convolution layer to convert features extracted by the backbone into vectorized keypoint representations, and two fully-connected layers to perform classification, respectively.

Inspired by label smoothing in traditional classification tasks~\cite{szegedy2016rethinking}, SimCC proposes a Gaussian label smoothing strategy that replaces the one-hot label with Gaussian distributed soft label centered at the ground-truth bin, which integrates the inductive bias in the model training and brings about significant performance improvement. We find this technique also coincides with the idea of SORD~\cite{sord} in the ordinal regression task. The soft label naturally encapsulates the rank likelihoods of the keypoint locations given the inter-class penalty distance defined by the label distribution.

\paragraph{Baseline}
We first remove the costly upsampling layers from the standard SimCC. Results in Table~\ref{tab:ablation:rm_upsampling} show that the trimmed SimCC has significantly lower complexity compared to the SimCC and heatmap-based baselines~\cite{xiao2018simple}, and still achieves promising accuracy. This indicates the efficiency of encoding global spatial information into disentangled one-dimension representations in localization tasks. By replacing the ResNet-50~\cite{he2016deep} backbone with the more compact CSPNext-m~\cite{lyu2022rtmdet}, we further reduce the model size and obtain a lightweight yet strong baseline, 69.7\% AP.

\begin{table}[htb]
\centering
\caption{Computational costs and accuracy of baseline methods. We show FLOPs and model parameters of prediction heads for a detailed comparison. ``SimCC*'' denotes the removal of upsampling layers from the standard SimCC head.}\label{tab:ablation:rm_upsampling}
\vspace{3pt}
\resizebox{0.45\textwidth}{!}{

\begin{tabular}{c|ccc}
\hline
                        & Heatmap       & SimCC    & SimCC*  \\ \hline
 Repr. Size    & 64$\times$48  & 512+384  & 512+384 \\
 AP                     & 71.8          & 72.1     & 71.3    \\
 Total FLOPs(G)         & 5.45          & 5.50     & 4.03    \\
 Total Params(M)        & 34.00         & 36.75    & 23.59   \\
 Head FLOPs(G)          & 1.425         & 1.472    & 0.002   \\
 Head Params(M)         & 10.492        & 13.245   & 0.079   \\  \hline
\end{tabular}}
\end{table}

\subsection{Training Techniques}\label{sec:training_techniques}

\paragraph{Pre-training}
Previous works~\cite{blazepose, li2021human} show that pre-training the backbone using the heatmap-based method can improve the model accuracy. We adopt UDP~\cite{Huang_2020_CVPR} method for the backbone pre-training. This improves the model from 69.7\% AP to 70.3\% AP. We use this technique as a default setting in the following sections.

\paragraph{Optimization Strategy}
We adopt the optimization strategy from~\cite{lyu2022rtmdet}. The Exponential Moving Average (EMA) is used for alleviating overfitting (70.3\% to 70.4\%). The flat cosine annealing strategy improves the accuracy to 70.7\% AP. We also inhibit weight decay on normalization layers and biases. 

\paragraph{Two-stage training augmentations}
Following the training strategy in~\cite{lyu2022rtmdet}, we use a strong-then-weak two-stage augmentation. First using strong data augmentations to train 180 epochs and then a weak strategy for 30 epochs. During the strong stage, we use a large random scaling range [0.6, 1.4], and a large random rotation factor, 80, and set the Cutout~\cite{TerranceDeVries2017ImprovedRO} probability to 1. According to AID~\cite{aid}, Cutout helps to prevent the model from overfitting to the image textures and encourages it to learn the pose structure information. In the weak strategy stage, we turn off the random shift, use a small random rotation range, and set the Cutout probability to 0.5 to allow the model to fine-tune in a domain that more closely matches the real image distribution. 

\subsection{Module Design}\label{sec:module}

\begin{figure*}[htbp]
    \includegraphics[width=\textwidth]{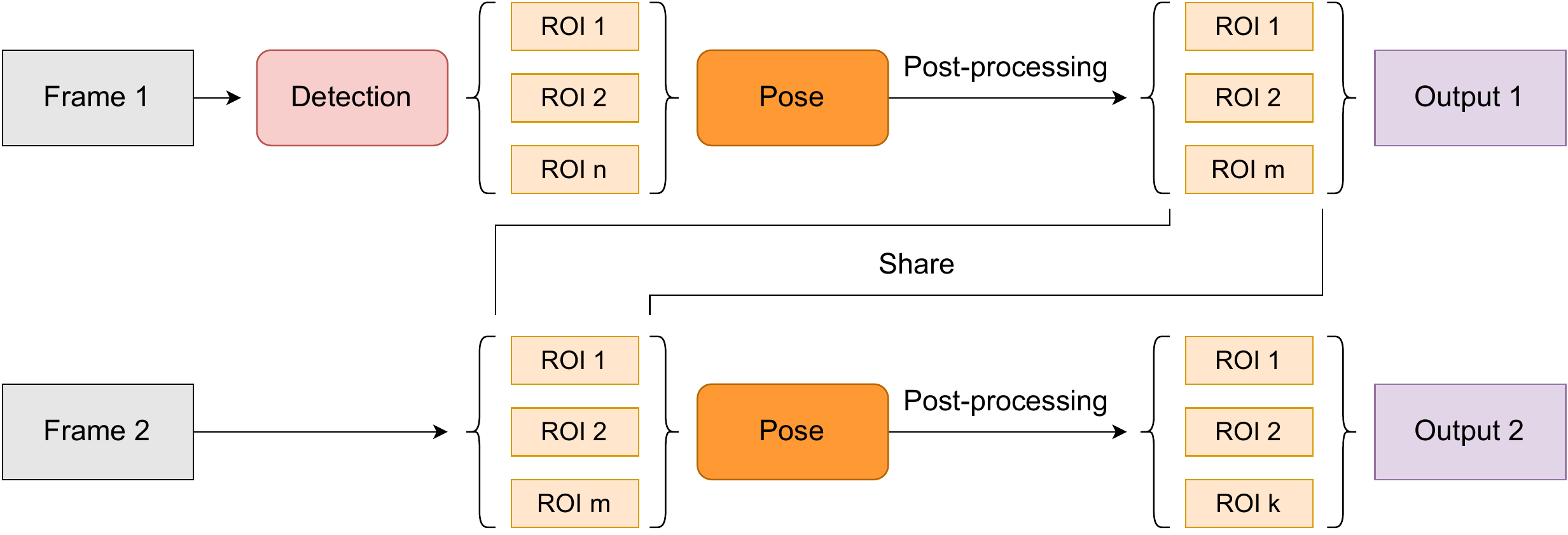}
    \caption{Inference pipeline of RTMPose.}
    \label{fig:pipeline}
\end{figure*}

\paragraph{Feature dimension}
We observe that the model performance increases along with higher feature resolution. Therefore, we use a fully connected layer to expand the 1D keypoint representations to a desired dimension controlled by the hyper-parameter. In this work, we use 256 dimensions and the accuracy is improved from 71.2\% AP to 71.4\% AP.

\paragraph{Self-attention module}
To further exploit the global and local spatial information, we refine the keypoint representations with a self-attention module, inspired by~\cite{li2021tokenpose, yang2021transpose}. We adopt the transformer variant, Gated Attention Unit (GAU)~\cite{Hua2022TransformerQI}, which has faster speed, lower memory cost, and better performance compared to the vanilla transformer~\cite{vaswani2017attention}. Specifically, GAU improves the Feed-Forward Networks (FFN) in the transformer layer with Gated Linear Unit (GLU)~\cite{glu}, and integrates the attention mechanism in an elegant form:

\begin{equation}\label{eq:gau}
\begin{split}
    U&=\phi_u(XW_{u}) \\
    V&=\phi_v(XW_{v}) \\
    O&=(U \odot AV)W_o
\end{split}
\end{equation}

where $\odot$ is the pairwise multiplication (Hadamard product) and $\phi$ is the activation function. In this work we implement the self-attention as follows:

\begin{equation}
A = \frac {1}{n} \mathit{relu} ^2( \frac {Q(X)K(Z)^\top}{ \sqrt {s} }), Z=\phi_z( X W_z)
\end{equation}

where $s = 128$, $Q$ and $K$ are simple linear transformations, and $\mathit{relu}^2(\cdot )$ is ReLU then squared. This self-attention module brings about a 0.5\% AP (71.9\%) improvement to the model performance.

\subsection{Micro Design}\label{sec:micro}

\paragraph{Loss function}
We treat the coordinate classification as an ordinal regression task and follow the soft label encoding proposed in SORD~\cite{sord}:

\begin{equation}\label{eq:sord}
y_i=\frac{e^{\phi(r_t, r_i)}}{\sum^K_{k=1}e^{\phi(r_t,r_k)}}
\end{equation}

where $\phi(r_t, r_i)$ is a metric loss function of our choice that penalizes how far the true metric value of $r_t$ is from the rank $r_i \in Y$. In this work, we adopt the unnormalized Gaussian distribution as the inter-class distance metric:

\begin{equation}
\phi(r_t, r_i)=e^{\frac{-(r_t-r_i)^2}{2\sigma^2}}
\end{equation}

Note that Eq.~\ref{eq:sord} can be seen as computing Softmax for all $\phi(r_t, r_i)$. We add temperatures in the Softmax operation for both model outputs and soft labels further adjust the normalized distribution shape:

\begin{equation}
y_i=\frac{e^{\phi(r_t, r_i)/\tau}}{\sum^K_{k=1}e^{\phi(r_t,r_l)/\tau}}
\end{equation}

According to the experimental results, using $\tau=0.1$ can improve the performance from 71.9\% to 72.7\%. 

\paragraph{Separate $\sigma$}
In SimCC, the horizontal and vertical labels are encoded using the same $\sigma$. We empirically explore a simple strategy to set separate $\sigma$ for them:

\begin{equation}
\sigma = \sqrt{\frac{W_{S}}{16}}
\end{equation}

where $W_{S}$ is the bin number in the horizontal and vertical directions respectively. This step improves the accuracy from 72.7\% to 72.8\%.

\paragraph{Larger convolution kernel}
We experiment with different kernel sizes of the last convolutional layer and find that using a larger kernel size gives a performance improvement over using $1\times1$ kernel. Finally, we chose to use a $7\times7$ convolutional layer, which achieves 73.3\% AP. We compare model performances with different kernel sizes in Table~\ref{tab:ablation:kernel_sizes}. Additionally, we also compare the effect of different temperature factors in Table~\ref{tab:ablation:temperature} using the final model architecture.

\begin{table}
    \centering
    \begin{minipage}[t]{0.45\linewidth}
        \centering
        \caption{Comparison of different kernel sizes}\label{tab:ablation:kernel_sizes}
        \vspace{6pt}
        \begin{tabular}{ll}
        \hline
        Kernel Size & mAP  \\ \hline
        1x1         & 72.8 \\
        3x3         & 73.2 \\
        5x5         & 73.2 \\
        \ours{7x7}  & \ours{73.3} \\
        9x9         & 73.0 \\
        \hline
        \end{tabular}
    \end{minipage}
    \hfill
    \begin{minipage}[t]{0.45\linewidth}
        \centering
        \caption{Comparison of different temperature factors.}\label{tab:ablation:temperature}
        \vspace{6pt}
        \begin{tabular}{cc}
        \hline
        1/$\tau$ & mAP  \\ \hline
        1         & unstable \\
        5         & 73.1 \\
        \ours{10} & \ours{73.3} \\
        15         & 73.0 \\
        \hline
        \end{tabular}
    \end{minipage}
\end{table}

\paragraph{More epochs and multi-dataset training}
Increasing the training epochs brings extra gains to the model performance. Specifically, training 270 and 420 epochs reach 73.5\% AP and 73.7\% AP respectively. To further exploit the model's potential, we enrich the training data by combining COCO~\cite{lin2014coco} and AI Challenger~\cite{JiahongWu2017AIC} datasets together for pre-training and fine-tuning, with a balanced sampling ratio. The performance finally achieves 75.3\% AP.

\subsection{Inference pipeline}
Beyond the pose estimation model, we further optimize the overall top-down inference pipeline for lower latency and better robustness. We use the skip-frame detection mechanism as in BlazePose~\cite{blazepose}, where human detection is performed every K frames, and in the interval frames the bounding boxes are generated from the last pose estimation results. Additionally, to achieve smooth prediction over frames, we use OKS-based pose NMS and OneEuro~\cite{casiez2012oneeuro} filter in the post-processing stage.

\section{Experiments}\label{sec:Experiments}

\begin{table*}[h]
	\begin{center}
 \caption{Body pose estimation results on COCO validation set. We only report GFLOPs of pose model and ignore the detection model. Flip test is not used.}
 
 \label{tab:compare_coco}
    \resizebox{\linewidth}{!}{
		\begin{tabular}{c|l|c|c|c|c|c|c|c}
		    \toprule
			\multicolumn{2}{c|}{Methods} 
            & Backbone & Detector & Det. Input Size & Pose Input Size & GFLOPs & AP & Extra Data  \\
			\midrule
			\multirow{6}{*}{PaddleDetection~\cite{ppdet2019}} 
            & TinyPose & Wider NLiteHRNet & YOLOv3 & $608 \times 608$ & $128 \times 96$ & 0.08 & 52.3  \\ 
			& TinyPose & Wider NLiteHRNet & YOLOv3 & $608 \times 608$ & $256 \times 192$ & 0.33 & 60.9  \\ 
            \cmidrule{2-8}
            & TinyPose & Wider NLiteHRNet & Faster-RCNN & N/A & $128 \times 96$ & 0.08 & 56.1 & AIC(220K) \\ 
			& TinyPose & Wider NLiteHRNet & Faster-RCNN & N/A & $256 \times 192$ & 0.33 & 65.6 &+Internal(unknown) \\ 
            \cmidrule{2-8}
			& TinyPose & Wider NLiteHRNet & PicoDet-s & $320 \times 320$ & $128 \times 96$ & 0.08 & 48.4  \\ 
			& TinyPose & Wider NLiteHRNet & PicoDet-s & $320 \times 320$ & $256 \times 192$ & 0.33 & 56.5  \\ 
            \midrule
            \multirow{6}{*}{AlphaPose~\cite{alphapose}} 
			& FastPose & ResNet 50 & YOLOv3 & $608 \times 608$ & $256 \times 192$ & 5.91 & 71.2 &\multirow{6}{*}{-} \\ 
			& FastPose(DUC) & ResNet-50 & YOLOv3 & $608 \times 608$ & $256 \times 192$ & 9.71 & 71.7 \\ 
			& FastPose(DUC) & ResNet-152 & YOLOv3 & $608 \times 608$ & $256 \times 192$ & 15.99 & 72.6 \\ 
            \cmidrule{2-8}
            & FastPose & ResNet 50 & Faster-RCNN & N/A & $256 \times 192$ & 5.91 & 69.7 \\ 
			& FastPose(DUC) & ResNet-50 & Faster-RCNN & N/A & $256 \times 192$ & 9.71 & 70.3 \\ 
			& FastPose(DUC) & ResNet-152 & Faster-RCNN & N/A & $256 \times 192$ & 15.99 & 71.3 \\ 
			\midrule
			\multirow{22}{*}{MMPose~\cite{mmpose2020}}
            & RTMPose-t & CSPNeXt-t & Faster-RCNN  & N/A & $256 \times 192$ & 0.36 & 65.8 &\multirow{4}{*}{-}\\ 
            & RTMPose-s & CSPNeXt-s & Faster-RCNN  & N/A & $256 \times 192$ & 0.68 & 69.6 \\ 
            & RTMPose-m & CSPNeXt-m & Faster-RCNN  & N/A & $256 \times 192$ & 1.93 & 73.6 \\ 
            & RTMPose-l & CSPNeXt-l & Faster-RCNN  & N/A & $256 \times 192$ & 4.16 & 74.8 \\ 
            \cmidrule{2-9}
            & RTMPose-t & CSPNeXt-t & YOLOv3 & $608 \times 608$ & $256 \times 192$ & 0.36 & 66.0&\multirow{18}{*}{AIC(220K)} \\ 
            & RTMPose-s & CSPNeXt-s & YOLOv3 & $608 \times 608$ & $256 \times 192$ & 0.68 & 70.3 \\ 
            & RTMPose-m & CSPNeXt-m & YOLOv3 & $608 \times 608$ & $256 \times 192$ & 1.93 & 74.7 \\ 
            & RTMPose-l & CSPNeXt-l & YOLOv3 & $608 \times 608$ & $256 \times 192$ & 4.16 & 75.7 \\ 
            \cmidrule{2-8}
            & RTMPose-t & CSPNeXt-t & Faster-RCNN & N/A & $256 \times 192$ & 0.36 & 67.1 \\ 
            & RTMPose-s & CSPNeXt-s & Faster-RCNN & N/A & $256 \times 192$ & 0.68 & 71.1 \\ 
            & RTMPose-m & CSPNeXt-m & Faster-RCNN & N/A & $256 \times 192$ & 1.93 & 75.3 \\ 
            & RTMPose-l & CSPNeXt-l & Faster-RCNN & N/A & $256 \times 192$ & 4.16 & 76.3 \\ 
            \cmidrule{2-8}
            & RTMPose-t & CSPNeXt-t & PicoDet-s & $320 \times 320$ & $256 \times 192$ & 0.36 & 64.3 \\ 
            & RTMPose-s & CSPNeXt-s & PicoDet-s & $320 \times 320$ & $256 \times 192$ & 0.68 & 68.8 \\ 
            & RTMPose-m & CSPNeXt-m & PicoDet-s & $320 \times 320$ & $256 \times 192$ & 1.93 & 73.2 \\ 
            & RTMPose-l & CSPNeXt-l & PicoDet-s & $320 \times 320$ & $256 \times 192$ & 4.16 & 74.2 \\ 
            \cmidrule{2-8}
            & RTMPose-t & CSPNeXt-t & RTMDet-nano & $320 \times 320$ & $256 \times 192$ & 0.36 & 64.4 \\ 
            & RTMPose-s & CSPNeXt-s & RTMDet-nano & $320 \times 320$ & $256 \times 192$ & 0.68 & 68.5 \\ 
            & RTMPose-m & CSPNeXt-m & RTMDet-nano & $320 \times 320$ & $256 \times 192$ & 1.93 & 73.2 \\ 
            & RTMPose-l & CSPNeXt-l & RTMDet-nano & $320 \times 320$ & $256 \times 192$ & 4.16 & 74.2 \\ 
            & RTMPose-m & CSPNeXt-m & RTMDet-m & $640 \times 640$ & $256 \times 192$ & 1.93 & 75.7 \\ 
            & RTMPose-l & CSPNeXt-l & RTMDet-m & $640 \times 640$ & $256 \times 192$ & 4.16 & 76.6 \\ 
			\bottomrule
		\end{tabular}
	}
	\end{center}
\end{table*}

\subsection{Settings}

The training settings in our experiments are shown in Tabel.~\ref{tab:train_setting}. As described in Sec.~\ref{sec:training_techniques}, we conduct a heatmap-based pre-training~\cite{Huang_2020_CVPR} which follows the same training strategies used in the fine-tuning except for shorter epochs. All our models are trained on 8 NVIDIA A100 GPUs. And we evaluate the model performance by mean Average Precision (AP).

\begin{table*}[h]
\begin{center}
 \caption{Body pose estimation results on COCO-SinglePerson validation set. We sum up top-down methods' GFLOPs of detection and pose for a fair comparison with bottom-up methods. ``*'' denotes double inference. Flip test is not used.}\label{tab:compare_coco_single}

    \resizebox{\linewidth}{!}{
		\begin{tabular}{c|l|c|c|c|c|c|c|c}
		    \toprule
			\multicolumn{2}{c|}{Methods} 
            & Backbone & Detector & Det. Input Size & Pose Input Size & GFLOPs & AP & Extra Data  \\
			\midrule
            \multirow{2}{*}{MediaPipe~\cite{blazepose}} 
            & BlazePose-Lite & BlazePose & N/A & $256 \times 256$ & N/A & N/A & 29.3  &\multirow{2}{*}{Internal(85K)}\\ 
			& BlazePose-Full & BlazePose & N/A & $256 \times 256$ & N/A & N/A & 35.4  \\ 
            \midrule
            \multirow{2}{*}{MoveNet~\cite{RonnyVotel2023NextGenerationPD}} 
            & Lightning & MobileNetv2 & N/A & $192 \times 192$ & N/A & 0.54 & 53.6*  &\multirow{2}{*}{Internal(23.5K)}\\ 
			& Thunder & MobileNetv2 depth$\times1.75$ & N/A & $256 \times 256$ & N/A & 2.44 &  64.8*  \\ 
            \midrule
            \multirow{2}{*}{PaddleDetection~\cite{ppdet2019}} 
            & TinyPose & Wider NLiteHRNet & PicoDet-s & $320 \times 320$ & $128 \times 96$ & 0.55 & 58.6 & AIC(220K) \\ 
		&  TinyPose & Wider NLiteHRNet & PicoDet-s & $320 \times 320$ & $256 \times 192$ & 0.80     & 69.4 &+Internal(unknown)  \\ 
            \midrule
			\multirow{4}{*}{MMPose~\cite{mmpose2020}}
			& RTMPose-t & CSPNeXt-t & RTMDet-nano & $320 \times 320$ & $256 \times 192$ & 0.67 & 72.1 &\multirow{4}{*}{AIC(220K)}\\ 
            & RTMPose-s & CSPNeXt-s & RTMDet-nano & $320 \times 320$ & $256 \times 192$ & 0.91 & 77.1 \\ 
            & RTMPose-m & CSPNeXt-m & RTMDet-nano & $320 \times 320$ & $256 \times 192$ & 2.23 & 82.4 \\ 
            & RTMPose-l & CSPNeXt-l & RTMDet-nano & $320 \times 320$ & $256 \times 192$ & 4.47 & 83.5 \\ 
			\bottomrule
		\end{tabular}
	}
\end{center}
\end{table*}

\begin{table*}[h]
\begin{center}
 \caption{Whole-body pose estimation results on COCO-WholeBody~\cite{zoomnet,xu2022zoomnas} V1.0 dataset. We only report the input size and GFLOPs of pose models in top-down approaches and ignore the detection model. ``*'' denotes the model is pre-trained on AIC+COCO. ``\dag'' indicates multi-scale testing. Flip test is used.}\label{tab:compare_cocowhole}
    \vspace{3pt}
    \resizebox{0.9\linewidth}{!}{
		\begin{tabular}{c|l|c|c|cc|cc|cc|cc|cc}
			\toprule
			  & Method & Input Size & GFLOPs & \multicolumn{2}{c|}{whole-body} & \multicolumn{2}{c|}{body}  & \multicolumn{2}{c|}{foot}  & \multicolumn{2}{c|}{face}  & \multicolumn{2}{c}{hand} \\
			\cmidrule{5-14}
			& &   &   &  AP     & AR     & AP   & AR     &  AP  & AR     & AP    & AR   &  AP     & AR  \\
			\midrule
			Whole- & SN\dag~\cite{hidalgo2019single} & N/A & 272.3 & 32.7 & 45.6 & 42.7 & 58.3 & 9.9 & 36.9 & 64.9 & 69.7 & 40.8 & 58.0  \\ 
            body & OpenPose~\cite{openpose} & N/A & 451.1 & 44.2 & 52.3 & 56.3 & 61.2 & 53.2 & 64.5 & 76.5 & 84.0 & 38.6 & 43.3  \\ 
            \midrule
			Bottom- & PAF\dag~\cite{cao2017realtime} & 512$\times$512 & 329.1 & 29.5 & 40.5 & 38.1 & 52.6 & 5.3 & 27.8 & 65.6 & 70.1 & 35.9 & 52.8  \\ 
			up & AE~\cite{newell2017associative} & 512$\times$512 & 212.4 & 44.0 & 54.5 & 58.0 & 66.1 & 57.7 & 72.5 & 58.8 & 65.4 & 48.1 & 57.4  \\
			\midrule
            & DeepPose~\cite{Toshev_2014_CVPR} & 384$\times$288 & 17.3 & 33.5 & 48.4 & 44.4 & 56.8 & 36.8 & 53.7 & 49.3 & 66.3 & 23.5 & 41.0 \\
            & SimpleBaseline~\cite{xiao2018simple} & 384$\times$288 & 20.4 & 57.3 & 67.1 & 66.6 & 74.7 & 63.5 & 76.3 & 73.2 & 81.2 & 53.7 & 64.7 \\
			& HRNet~\cite{SunXLW19}  & 384$\times$288 & 16.0 & 58.6 & 67.4 & 70.1 & 77.3 & 58.6 & 69.2 & 72.7 & 78.3 & 51.6 & 60.4  \\
            & PVT~\cite{WenhaiWang2021PyramidVT} & 384$\times$288 & 19.7 & 58.9 & 68.9 & 67.3 & 76.1 & 66.0 & 79.4 & 74.5 & 82.2 & 54.5 & 65.4 \\
		Top- & FastPose50-dcn-si~\cite{alphapose} &  256$\times$192 & 6.1 & 59.2 & 66.5 & 70.6      & 75.6 & 70.2 & 77.5 & 77.5 & 82.5 & 45.7 & 53.9 \\
            down & ZoomNet~\cite{zoomnet} & 384$\times$288 & 28.5 & 63.0 & 74.2 & \textbf{74.5} &   \textbf{81.0} & 60.9 & 70.8 & 88.0 & 92.4 & 57.9 & 73.4   \\
            & ZoomNAS~\cite{xu2022zoomnas} & 384$\times$288 & 18.0 & \textbf{65.4} & \textbf{74.4} & 74.0 & 80.7 & 61.7 & 71.8 & 88.9 & \textbf{93.0} & \textbf{62.5} & \textbf{74.0} \\
            \cmidrule{2-14}
            & \ours{RTMPose-m*} & \ours{256$\times$192} & \ours{2.2} & \ours{58.2} & \ours{67.4} & \ours{67.3} & \ours{75.0} & \ours{61.5} & \ours{75.2} & \ours{81.3} & \ours{87.1} & \ours{47.5} & \ours{58.9}  \\
            & \ours{RTMPose-l*} & \ours{256$\times$192} & \ours{4.5} & \ours{61.1} & \ours{70.0} & \ours{69.5} & \ours{76.9} & \ours{65.8} & \ours{78.5} & \ours{83.3} & \ours{88.7} & \ours{51.9} & \ours{62.8}  \\
            & \ours{RTMPose-l*} & \ours{384$\times$288} & \ours{10.1} & \ours{64.8} & \ours{73.0} & \ours{71.2} & \ours{78.1} & \ours{\textbf{69.3}} & \ours{\textbf{81.1}} & \ours{88.2} & \ours{91.9} & \ours{57.9} & \ours{67.7} \\
            & \ours{RTMPose-x} & \ours{384$\times$288} & \ours{18.1} & \ours{65.2} & \ours{73.2} & \ours{71.2} & \ours{78.0} & \ours{68.1} & \ours{80.4} & \ours{\textbf{89.0}} & \ours{92.2} & \ours{59.3} & \ours{68.7} \\
            & \ours{RTMPose-x*} & \ours{384$\times$288} & \ours{18.1} & \ours{65.3} & \ours{73.3} & \ours{71.4} & \ours{78.4} & \ours{69.2} & \ours{81.0} & \ours{88.9} & \ours{92.3} & \ours{59.0} & \ours{68.5} \\
			\bottomrule
		\end{tabular}}
	\end{center}
\end{table*}

\begin{table}[h]
\small
\centering
\caption{Training settings for RTMPose models.}\label{tab:train_setting}
\vspace{3pt}
\begin{tabular}{l c}
\hline
optimizer & AdamW \\
base learning rate & 0.004 \\
learning rate schedule & Flat-Cosine \\
batch size & 1024 \\
warm-up iterations & 1000 \\
\hline
\multirow{2}{0.15\textwidth}{weight decay} & 0.05 (RTMPose-m/l)\\
& 0 (RTMPose-t/s)\\
\hline
\multirow{2}{0.15\textwidth}{EMA decay} & 0.9998 (RTMPose-s/m/l)\\
& no EMA (RTMPose-t)\\
\hline
\multirow{2}{0.15\textwidth}{training epochs} & 210 (pre-train)\\
& 420 (fine-tune)\\
\hline
\end{tabular}
\vspace{-5pt}
\end{table}

\subsection{Benchmark Results}

\begin{figure*}[htp]
    \centering
    \begin{minipage}[t]{0.48\textwidth}
    \centering
    \includegraphics[width=9cm]{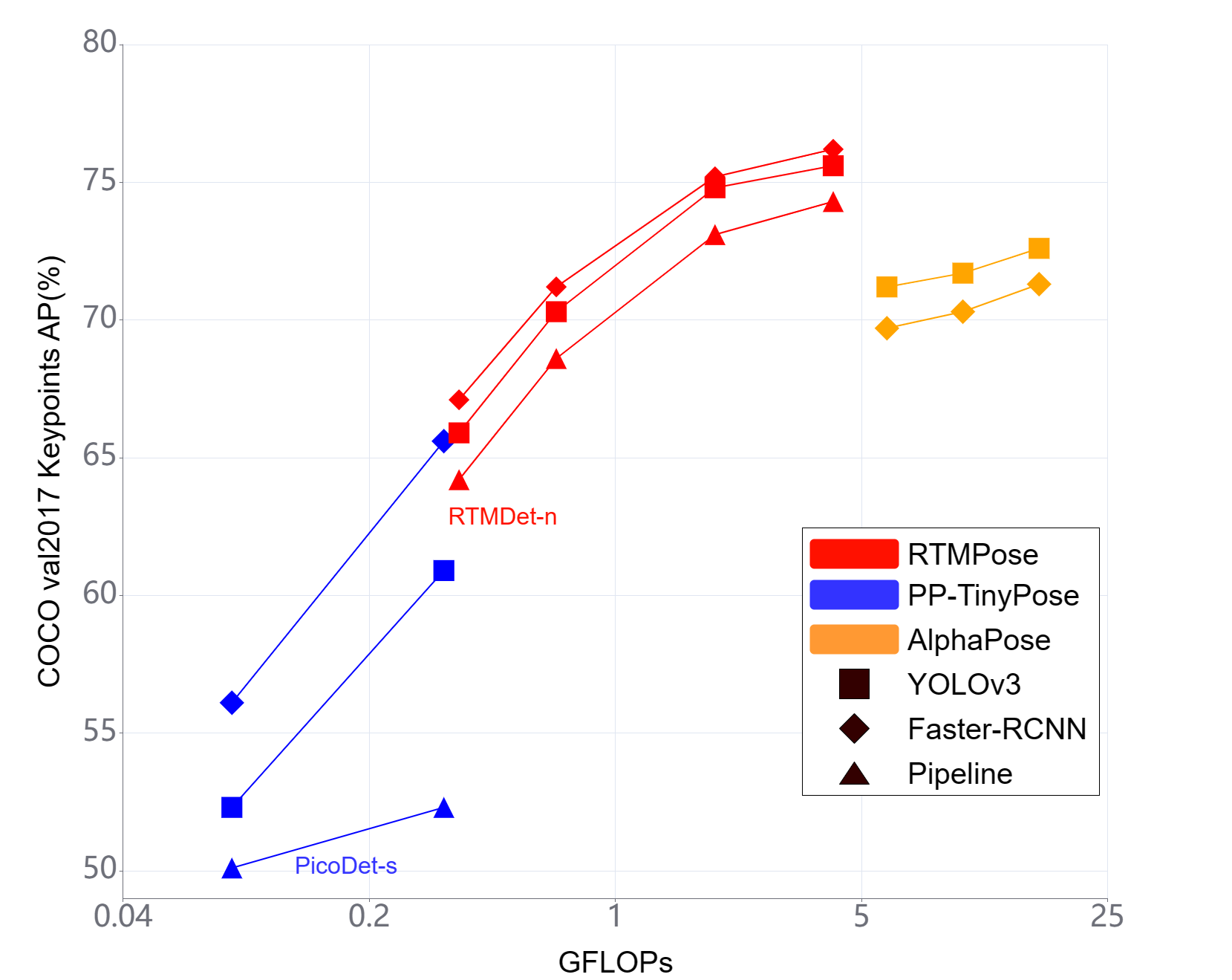}
    \end{minipage}
    \begin{minipage}[t]{0.48\textwidth}
    \centering
    \includegraphics[width=9cm]{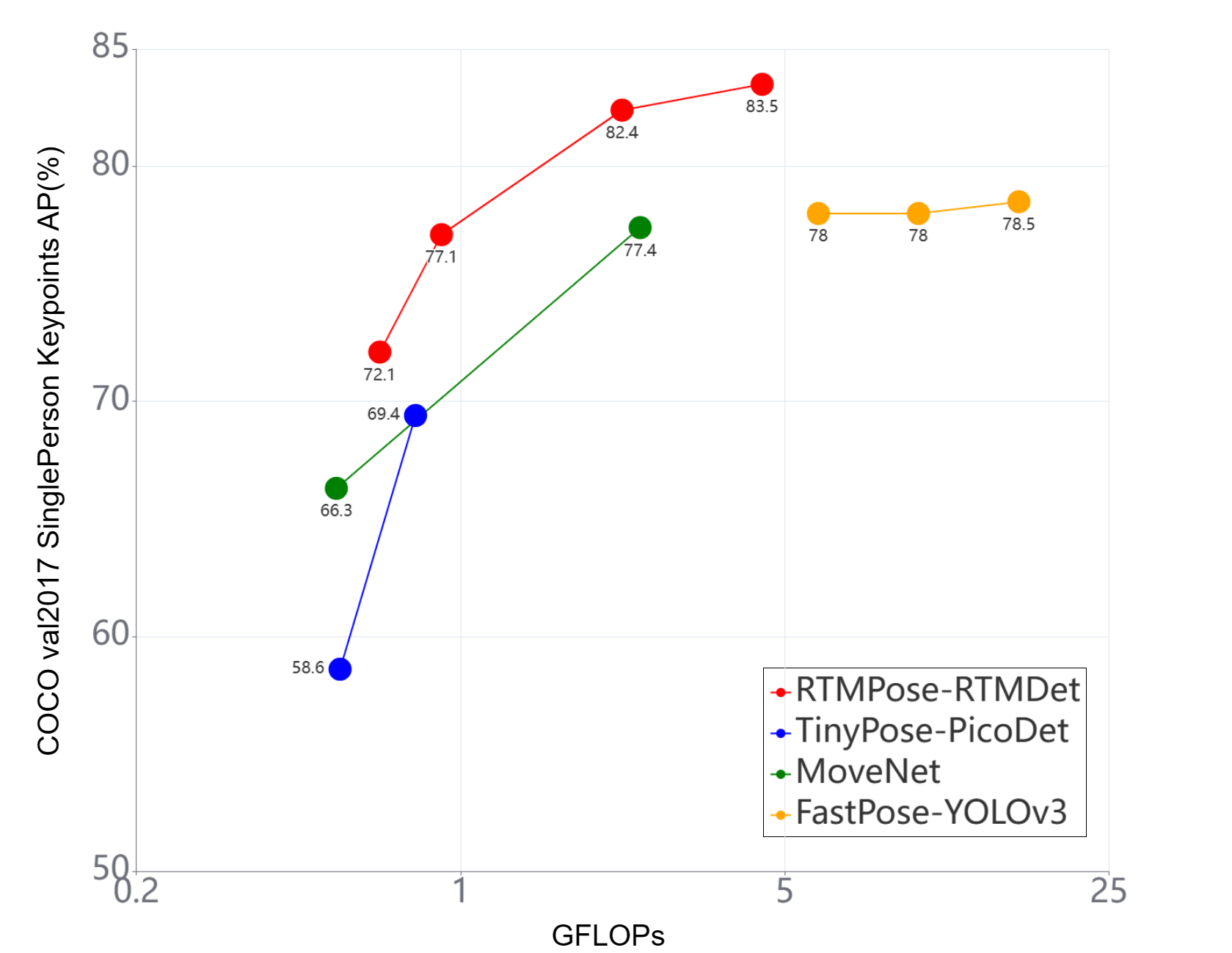}
    \end{minipage}
    \caption{Comparison of GFLOPs and accuracy. Left: Comparison of RTMPose and other open-source pose estimation libraries on full COCO val set. Right: Comparison of RTMPose and other open-source pose estimation libraries on COCO-SinglePerson val set.}
    \label{fig:compare_gflops}
\end{figure*}

\paragraph{COCO}
COCO~\cite{lin2014coco} is the most popular benchmark for 2d body pose estimation. We follow the standard splitting of \texttt{train2017} and \texttt{val2017}, which contains 118K and 5k images for training and validation respectively. We extensively study the pose estimation performance with different off-the-shelf detectors including YOLOv3~\cite{JosephRedmon2018YOLOv3AI}, Faster-RCNN~\cite{ShaoqingRen2015FasterRT}, and RTMDet~\cite{lyu2022rtmdet}. To conduct a fair comparison with AlphaPose~\cite{alphapose} which doesn't use extra training data, we also report the performance of RTMPose only trained on COCO. As shown in Table~\ref{tab:compare_coco}, RTMPose outperforms competitors by a large margin with much lower complexity and shows strong robustness for detection.

\paragraph{COCO-SinglePerson}
Popular pose estimation open-source algorithms like BlazePose~\cite{blazepose}, MoveNet~\cite{RonnyVotel2023NextGenerationPD}, and PaddleDetection~\cite{ppdet2019} are designed primarily for single-person or sparse scenarios, which are practical in mobile applications and human-machine interactions. For a fair comparison, we construct a COCO-SinglePerson dataset that contains 1045 single-person images from the COCO \texttt{val2017} set to evaluate RTMPose as well as other approaches. For MoveNet, we follow the official inference pipeline to apply a cropping algorithm, namely using the coarse pose prediction of the first inference to crop the input image and performing a second inference for better pose estimation results. The evaluation results in Table~\ref{tab:compare_coco_single} show that RTMPose archives superior performance and efficiency even compared to previous solutions tailored for the single-person scenario.

\paragraph{COCO-WholeBody}
We also validate the proposed RTMPose model on the whole-body pose estimation task with COCO-WholeBody~\cite{zoomnet, xu2022zoomnas} V1.0 dataset. As shown in Table~\ref{tab:compare_cocowhole}, RTMPose achieves superior performance and well balances accuracy and complexity. Specifically, our RTMPose-m model outperforms previous open-source libraries~\cite{openpose, alphapose, Yulitehrnet21} with significantly lower GFLOPs. And by increasing the input resolution and training data we obtain competitive accuracy with SOTA approaches~\cite{zoomnet, xu2022zoomnas}.

\paragraph{Other Datasets}
As shown in Table~\ref{tab:compare_others} and Table~\ref{tab:compare_mpii}, we further evaluate RTMPose on AP-10K~\cite{HangYu2021AP10KAB}, CrowdPose~\cite{JiefengLi2018CrowdPoseEC} and MPII~\cite{MykhayloAndriluka20142DHP} datasets. We report the model performance using ImageNet~\cite{JiaDeng2009ImageNetAL} pre-training for a fair comparison with baselines. Besides we also report the performance of our models pre-trained using a combination of COCO~\cite{lin2014coco} and AI Challenger (AIC)~\cite{JiahongWu2017AIC}, which achieves higher accuracy and can be easily reproduced by users with our provided pre-trained weights.

\subsection{Inference Speed}
We perform the export, deployment, inference, and testing of models by MMDeploy~\cite{mmdeploy} to test the inference speed on CPU and GPU respectively. Table~\ref{tab:ncnn_speed} demonstrates the comparison of inference speed on the mobile device. We deploy RTMPose on the Snapdragon 865 chip with ncnn and inference with 4 threads. The TensorRT inference latency is tested in the half-precision floating-point format (FP16) on an NVIDIA GeForce GTX 1660 Ti GPU, and the ONNX latency is tested on an Intel I7-11700 CPU with ONNXRuntime with 1 thread. The inference batch size is 1. All models are tested on the same devices with 50 times warmup and 200 times inference for fair comparison. For TinyPose~\cite{ppdet2019}, we test it with both MMDeploy and FastDeploy, and note that ONNXRuntime speed on MMDeploy is slightly faster  (10.58 ms vs. 12.84 ms). The results are shown in Table~\ref{tab:onnx_trt_speed} and Table~\ref{tab:pipeline_speed}. 

\begin{table}[h]
	\begin{center}
 \caption{Performance on different datasets. ``*'' denotes the model is pre-trained on AIC+COCO and fine-tuned on the corresponding dataset. Flip test is used.}\label{tab:compare_others}
  \vspace{3pt}
    \resizebox{\linewidth}{!}{
		\begin{tabular}{c|l|c|c|c|c}
		    \toprule
			Dataset & Methods & Backbone & Input Size & GFLOPs & AP  \\
            \midrule
			\multirow{4}{*}{AP-10K~\cite{HangYu2021AP10KAB}} 
            & SimpleBaseline~\cite{xiao2018simple} & ResNet-50 & $256 \times 256$ & 7.28 & 68.0 \\ 
			& HRNet~\cite{SunXLW19} & HRNet-w32 & $256 \times 256$ & 10.27 & 72.2 \\ 
            & \ours{RTMPose-m} & \ours{CSPNeXt-m} & \ours{$256 \times 256$} & \ours{\textbf{2.57}} & \ours{\textbf{68.4}}    \\ 
            & \ours{RTMPose-m*} & \ours{CSPNeXt-m} & \ours{$256 \times 256$} & \ours{\textbf{2.57}} & \ours{\textbf{72.2}}   \\ 
            \midrule
            \multirow{4}{*}{CrowdPose~\cite{JiefengLi2018CrowdPoseEC}} 
            & SimpleBaseline~\cite{xiao2018simple} & ResNet-50 & $256 \times 192$ & 5.46 & 63.7 \\ 
			& HRNet~\cite{SunXLW19} & HRNet-w32 & $256 \times 192$ & 7.7 & 67.5 \\ 
			& \ours{RTMPose-m} & \ours{CSPNeXt-m} & \ours{$256 \times 192$} & \ours{\textbf{1.93}} & \ours{\textbf{66.9}}  \\  
            & \ours{RTMPose-m*} & \ours{CSPNeXt-m} & \ours{$256 \times 192$} & \ours{\textbf{1.93}} & \ours{\textbf{70.6}} \\  
            \bottomrule
		\end{tabular}
	}
	\end{center}
\end{table}

\begin{table}[ht!]
	\begin{center}
 \caption{Comparison on MPII~\cite{MykhayloAndriluka20142DHP} validation set. ``*'' denotes the model is pre-trained on AIC+COCO and fine-tuned on MPII. Flip test is used.}\label{tab:compare_mpii}
  \vspace{3pt}
    \resizebox{\linewidth}{!}{
		\begin{tabular}{c|l|c|c|c|c}
		    \toprule
			Dataset & Methods & Backbone & Input Size & GFLOPs & PCKh@0.5  \\
            \midrule
			\multirow{6}{*}{MPII~\cite{MykhayloAndriluka20142DHP}} 
            & SimpleBaseline~\cite{xiao2018simple} & ResNet-50 & $256 \times 256$ & 7.28 & 88.2\\ 
			& HRNet~\cite{SunXLW19} & HRNet-w32 & $256 \times 256$ & 10.27 & 90.0 \\ 
            & SimCC~\cite{SimCC} & HRNet-w32 & $256 \times 256$ & 10.34 & 90.0 \\
            & TokenPose~\cite{li2021tokenpose} & L/D24 & $256 \times 256$ & 11.0 & 90.2 \\
			& \ours{RTMPose-m} & \ours{CSPNeXt-m} & \ours{$256 \times 256$} & \ours{\textbf{2.57}} & \ours{\textbf{88.9}}    \\ 
            & \ours{RTMPose-m*} & \ours{CSPNeXt-m} & \ours{$256 \times 256$} & \ours{\textbf{2.57}} & \ours{\textbf{90.7}}   \\ 
			\bottomrule
		\end{tabular}
	}
	\end{center}
\end{table}

\begin{table}[h]
\caption{Comparison of inference speed on Snapdragon 865. RTMPose models are deployed and tested using ncnn.}\label{tab:ncnn_speed}
 \vspace{-18pt}
	\begin{center}
    \resizebox{\linewidth}{!}{
		\begin{tabular}{c|l|c|c|c|c|c}
		    \toprule
			\multicolumn{2}{c|}{Methods} 
            & Input Size & GFLOPs & AP(GT) & FP32(ms) & FP16(ms)  \\
			\midrule
			\multirow{2}{*}{PaddleDetection~\cite{ppdet2019}} 
            & TinyPose & $128 \times 96$ & 0.08 & 58.4 & 4.57 & 3.27\\ 
			&  TinyPose & $256 \times 192$ & 0.33 & 68.3 & 14.07 & 8.33 \\ 
            \midrule
			\multirow{4}{*}{MMPose~\cite{mmpose2020}}
			& RTMPose-t & $256 \times 192$ & 0.36 & 68.4 & 15.84 & 9.02 \\ 
            & RTMPose-s & $256 \times 192$ & 0.68 & 72.8 & 25.01 & 13.89 \\ 
            & RTMPose-m & $256 \times 192$ & 1.93& 77.3 & 49.46 & 26.44\\ 
            & RTMPose-l & $256 \times 192$ & 4.16 & 78.3 & 85.75 & 45.37 \\ 
			\bottomrule
		\end{tabular}}
	\end{center}
\end{table}

\begin{table}[h]
\caption{Inference speed on CPU and GPU. RTMPose models are deployed and tested using ONNXRuntime and TensorRT respectively. Flip test is not used in this table. }\label{tab:onnx_trt_speed}
 \vspace{-3pt}
	\begin{center}
    \resizebox{\linewidth}{!}{
		\begin{tabular}{c|l|c|c|c|c|cc}
		    \toprule
			\multicolumn{2}{c|}{Results} 
            & Input Size & GFLOPs & AP & CPU(ms) & GPU(ms)  \\
			\midrule
			\multirow{7}{*}{COCO~\cite{lin2014coco}}
            & TinyPose & $256 \times 192$ & 0.33 & 65.6 & 10.580 & 3.055\\ 
            & LiteHRNet-30 & $256 \times 192$ & 0.42 & 66.3 & 22.750 & 6.561\\ 
			& \ours{RTMPose-t} & \ours{$256 \times 192$} & \ours{\textbf{0.36}} & \ours{\textbf{67.1}} & \ours{\textbf{3.204}} & \ours{\textbf{1.064}}\\ 
            & \ours{RTMPose-s} & \ours{$256 \times 192$} & \ours{\textbf{0.68}} & \ours{\textbf{71.2}} & \ours{\textbf{4.481}} & \ours{\textbf{1.392}}\\ 
            \cmidrule{2-7}
            & HRNet-w32+UDP & $256 \times 192$ & 7.7 & 75.1 & 37.734 & 5.133\\ 
            & \ours{RTMPose-m} & \ours{$256 \times 192$} & \ours{\textbf{1.93}}& \ours{\textbf{75.3}} & \ours{\textbf{11.060}} & \ours{\textbf{2.288}}\\ 
            & \ours{RTMPose-l} & \ours{$256 \times 192$} & \ours{\textbf{4.16}} & \ours{\textbf{76.3}} & \ours{\textbf{18.847}} & \ours{\textbf{3.459}}\\ 
            \midrule
			\multirow{5}{*}{COCO-}
             & HRNet-w32+DARK & $256 \times 192$ & 7.72& 57.8 & 39.051 & 5.154\\ 
            & \ours{RTMPose-m} & \ours{$256 \times 192$} & \ours{\textbf{2.22}}& \ours{\textbf{59.1}} & \ours{\textbf{13.496}} & \ours{\textbf{4.000}}\\ 
            & \ours{RTMPose-l} & \ours{$256 \times 192$} & \ours{\textbf{4.52}} & \ours{\textbf{62.2}} & \ours{\textbf{23.410}} & \ours{\textbf{5.673}}\\ 
            \cmidrule{2-7}
            WholeBody~\cite{zoomnet}& HRNet-w48+DARK & $384 \times 288$ & 35.52 & 65.3 & 150.765 & 13.974\\ 
            & \ours{RTMPose-l} & \ours{$384 \times 288$} & \ours{\textbf{10.07}} & \ours{\textbf{66.1}} & \ours{\textbf{44.581}}  & \ours{\textbf{7.678}}\\ 
			\bottomrule
		\end{tabular}}
	\end{center}
\end{table}

\begin{table}[h]
\caption{Pipeline Inference speed on CPU, GPU and Mobile device.}\label{tab:pipeline_speed}
	\begin{center}
    \resizebox{\linewidth}{!}{
		\begin{tabular}{c|l|c|c|c|c|c}
		    \toprule
            Model & Input Size & GFLOPs & Pipeline AP & CPU(ms) & GPU(ms) & Mobile(ms) \\
			\midrule
            RTMDet-nano & $320\times 320$ & 0.31 & \multirow{2}{*}{64.4} & \multirow{2}{*}{12.403} &  \multirow{2}{*}{2.467} &  \multirow{2}{*}{18.780}\\ 
            RTMPose-t & $256 \times 192$ & 0.36 &  &  & \\ 
            \midrule
            RTMDet-nano & $320\times 320$ & 0.31 & \multirow{2}{*}{68.5} & \multirow{2}{*}{16.658} &  \multirow{2}{*}{2.730} &  \multirow{2}{*}{21.683}\\ 
            RTMPose-s & $256 \times 192$ & 0.42 &  &  & \\ 
            \midrule
            RTMDet-nano & $320\times 320$ & 0.31 & \multirow{2}{*}{73.2} & \multirow{2}{*}{26.613} &  \multirow{2}{*}{4.312} &  \multirow{2}{*}{32.122}\\ 
            RTMPose-m & $256 \times 192$ & 1.93 &  &  & \\ 
            \midrule
            RTMDet-nano & $320\times 320$ & 0.31 & \multirow{2}{*}{74.2} & \multirow{2}{*}{36.311} &  \multirow{2}{*}{4.644} &  \multirow{2}{*}{47.642}\\ 
            RTMPose-l & $256 \times 192$ & 4.16 &  &  & \\ 
            \bottomrule
		\end{tabular}}
	\end{center}
\end{table}
\section{Conclusion}\label{sec:Conclusion}

This paper empirically explores key factors in pose estimation such as the paradigm, model architecture, training strategy, and deployment. Based on the findings we present a high-performance real-time multi-person pose estimation framework, RTMPose, which achieves excellence in balancing model performance and complexity and can be deployed on various devices (CPU, GPU, and mobile devices) for real-time inference. We hope that the proposed algorithm alone with its open-sourced implementation can meet some of the demand for applicable pose estimation in industry, and benefit future explorations on the human pose estimation task.

{\small
\bibliographystyle{ieee_fullname}
\bibliography{egbib}
}

\end{document}